\documentclass[letterpaper, 10pt, conference]{ieeeconf}      % Use this line for a4 paper

\IEEEoverridecommandlockouts                              % This command is only needed if 
\usepackage{times} % font face
\usepackage{graphicx} % for \includegraphics
\usepackage{bm} % for \bm
\usepackage{amsmath} % for align environment
\usepackage{amssymb} % for \mathbb
\usepackage{color} % for \textcolor
\usepackage[ruled,linesnumbered]{algorithm2e} % for pseudocode
\usepackage[short]{optidef} % for argmini environment
\usepackage{gensymb} %for generic symbols
\usepackage{multirow}
\usepackage[mathscr]{eucal}

\usepackage{xparse,soul} % for \hl and \st
\soulregister\cite7 % fix \cite
\soulregister\citep7 % fix \citep
\soulregister\citet7 % fix \citet
\soulregister\ref7 % fix \ref
\soulregister\pageref7 % fix \pageref

\usepackage[flushleft]{threeparttable}
\usepackage{makecell}

\usepackage{fancyhdr}
\fancypagestyle{firstpage}{
    \chead{This paper has been accepted for publication at the 2025 International Conference on Intelligent Robots and Systems.}}
\usepackage{cite}

\title{Towards Design and Development of a Concentric Tube Steerable Drilling Robot for Creating S-shape Tunnels for  Pelvic Fixation Procedures}
\author{Yash Kulkarni$^{*1}$, Susheela Sharma$^{*1}$, Sarah Go$^{2}$, Jordan P. Amadio$^{3}$, Mohsen Khadem$^{4}$, \\ and Farshid Alambeigi$^{1}$ \IEEEmembership{Member, IEEE}
\thanks{*These authors contributed equally to this work.}
\thanks{**This work was supported in part by the National Institute Of Biomedical Imaging and Bioengineering of the National Institutes of Health under Award Number R21EB030796 and in part by the Collaborative Accelerator for Transformative Research Endeavors grant, jointly awarded by The University of Texas at Austin and The University of Texas MD Anderson Cancer Center.}
\thanks{$^{1}$Y.~Kulkarni, S.~Sharma, and F.~Alambeigi are with the Walker Department of Mechanical Engineering and Texas Robotics at the University of Texas at Austin, Austin, TX, 78712, USA. Email: \{kulkarni.yash08,sheela.sharma\}@utexas.edu, farshid.alambeigi@austin.utexas.edu}.
\thanks{$^{2}$S.~Go is with the Chandra Department of Electrical and Computer Engineering at The University of Texas at Austin, Austin, TX, 78712, USA. Email: sarah.go@utexas.edu}
\thanks{$^{3}$J.~P.~ Amadio is with the Department of Neurosurgery, The University of Texas Dell Medical School, TX, 78712.}
\thanks{$^{4}$M.~Khadem is with the School of Informatics, University of Edinburgh, UK. }}

\usepackage{color} % for \textcolor

\begin{document}
\maketitle
\thispagestyle{firstpage}
\pagestyle{empty}
		
	%%%%%%%%%%%%%%%%%%%%%%%%%%%%%%%%%%%%%%%%%%%%%%%%%%%%%%%%%%%%%%%%%%%%%%%%%%%%%%%%
\begin{abstract}
Current pelvic fixation techniques rely on rigid drilling tools, which inherently constrain the placement of rigid medical screws in the complex anatomy of pelvis. These constraints prevent medical screws from following anatomically optimal pathways and force clinicians to fixate screws in linear trajectories. This suboptimal approach, combined with the unnatural placement of the excessively long screws, lead to complications such as screw misplacement, extended surgery times, and increased radiation exposure due to repeated X-ray images taken ensure to safety of procedure. To address these
challenges, in this paper, we present the  design and development of a unique 4-degree-of-freedom (DoF) pelvic concentric tube steerable drilling robot (\textit{pelvic} CT-SDR). The \textit{pelvic} CT-SDR is capable of creating long S-shaped drilling trajectories that follow the natural curvatures of the pelvic anatomy. The performance of the  \textit{pelvic} CT-SDR was thoroughly evaluated through several S-shape drilling experiments in simulated bone phantoms.
 
\end{abstract}

%%%%%%%%%%%%%%%%%%%%%%%%%%%%%%%%%%%%%%%%%%%%%%%%%%%%%%%%%%%%%%%%%%%%%%%%%%%%%%%%
\section{Introduction}
\begin{figure}[t]
    \centering
    \includegraphics[width=0.9\columnwidth]{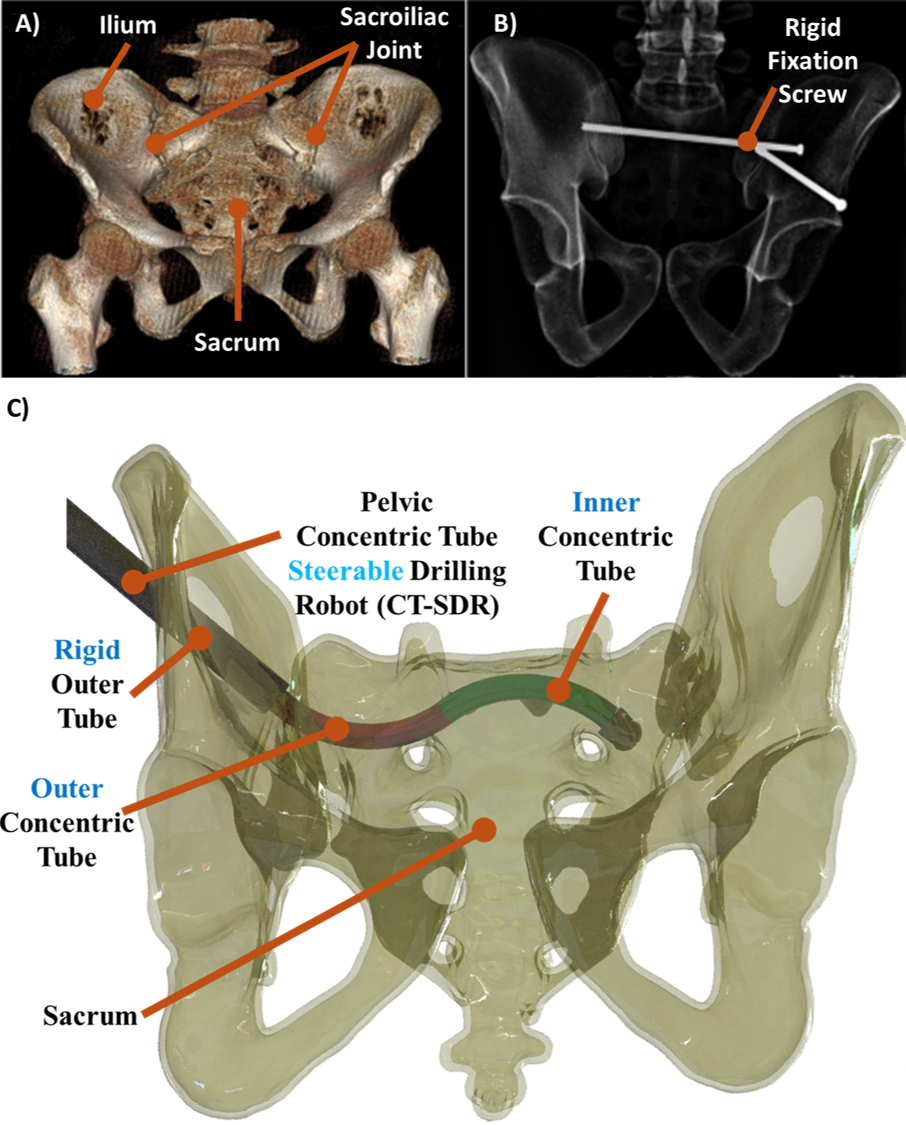}
     \caption{(A) CT scan of pelvis with anatomical components labeled \cite{Li2022TheEO}. (B) X-ray images of pelvis with conventional rigid screws fixated in different configuration\cite{Carlson2022PlacementOL}. (C) Conceptual illustration of proposed  4 degree of freedom \textit{pelvic} CT-SDR. }
    \label{fig:Anat}
\end{figure}

Pelvic fractures are complex musculoskeletal injuries that account for 3\% of all fractures \cite{Grotz2005OpenPF} with their incidences predicted to exceed 3 million cases in the United States by 2025 \cite{Hu2022EpidemiologyAB}. Pelvic fractures, commonly caused by high velocity incidents such as motor vehicle collisions \cite{Alwaal2014TheIC}, have a high disability rate of 60\% \cite{Burkhardt2014PelvicFI,Hermans2018OpenPF,Zhao2022DesignAE} with mortality rates ranging from 9\%-50\% primarily due to associated hemorrhaging \cite{Chong1997PelvicFA}. While non-surgical methods are sometimes recommended for minor pelvic fractures, sacral fractures and sacroiliac joint dislocations often require percutaneous sacroiliac screw fixation\cite{Bishop2012OsseousFP,Yang2022ComputeraidedAP} (see Fig. \ref{fig:Anat}-A and Fig. \ref{fig:Anat}-B). A critical component of this fixation method is  to first insert a guide wire through the pelvis to establish the desired drilling trajectory.  A cannulated rigid drilling instrument is then advanced over the guide wire, ensuring precise and controlled drilling over the guide wire. As shown in Fig. \ref{fig:Anat}, these trajectories serve as pathways for long and rigid fixation screws, which are then fixed in place to stabilize the fractured pelvis for restoring functionality and mobility to patients.

 Despite the success of percutaneous sacroiliac screw fixation in improving patient outcomes, screw misplacement remains a critical concern, as it can lead to serious complications such as neurovascular injuries \cite{Gilani2023TheUO}. The causes of screw misplacement are multi-factorial, ranging from long length and rigidity of the used screws and drilling instruments,  reliance on 2D imaging for screw placement  to inadequate preoperative planning \cite{Routt1997IliosacralSC,Gilani2023TheUO}. In particular, Yang et al. \cite{Yang2022ComputeraidedAP} have highlighted that the complex anatomy of the pelvis is the main reason making it difficult to identify a suitable fixation path using  rigid drilling and fixation instruments. Figure \ref{fig:Anat} highlights the complex anatomy of the pelvis with multiple changing curvatures and variable thickness along with vital nerves surrounding the pelvis.

\begin{figure*}[t]
    \centering
    \includegraphics[width=1\linewidth]{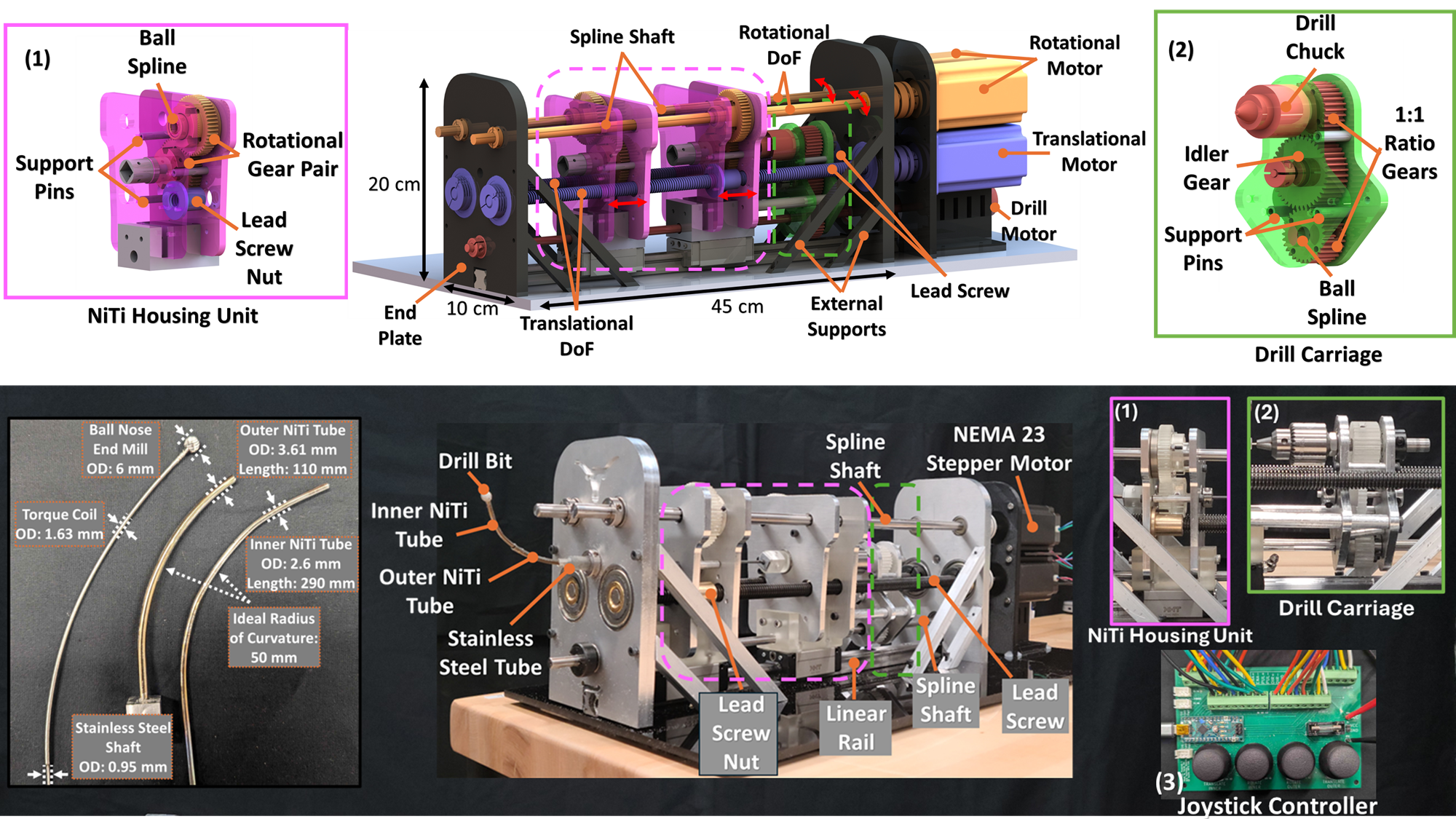}
     \caption{Overview of the pelvic CT-SDR and its key components. Top: schematic diagrams of the system, including (1) the NiTi Housing Unit for rotational and translational control of the NiTi tubes, and (2) the Drill Carriage which controls rotational motion of the drill tip. Bottom: Physical prototype with labeled components, including the drill bit, NiTi tubes, stainless steel tube, stepper motor, and lead screw. Detailed views of the NiTi Housing Unit, Drill Carriage, and Joystick Controller are also provided.}
    \label{fig:Drill}
\end{figure*}

To address the above-mentioned limitations, researchers have begun exploring the use of robots \cite{Gilani2023TheUO} and augmented reality navigation \cite{Zang2024AugmentedRN} to assist surgeons in accurate drilling and placement of the existing long and rigid screws across the pelvic anatomy  \cite{Zang2024AugmentedRN}.
Moreover, towards considering the complex anatomy of pelvis and updating the existing fixation methods, Yang et al. \cite{Yang2022ComputeraidedAP} proposed the use of an arc fixation channel to enhance screw placement. They developed a path planning algorithm to find the optimal arc trajectory and utilized simulation studies  through finite element analysis to assess the biomechanical impact between an optimally picked arc fixation screws and conventional long and rigid screws.
While this simulation study demonstrates that  development of a system capable of creating arc channels could be beneficial for pelvic fixation, the work doesn't provide any real validation regarding their proposed device \cite{Yang2022ComputeraidedAP}. Similarly, industrial products have recently  been developed based on the idea of curved fixation such as the CurvaFix IM Implant (CurvaFix, Bellevue, WA, USA) \cite{Miller2022CurvafixAN}. Nevertheless, these products still utilize the existing rigid instruments for drilling and creating the insertion tunnel for their flexible implant. 

Towards addressing challenges in creating nonlinear drilling trajectories for implanting flexible implants, researchers have recently developed different types of  steerable drilling robots for creating nonlinear trajectories for orthopedics and neurosurgical  procedures \cite{Sharma2023ISMR,Sharma2024TBME,alambeigi2019use,alambeigi2017curved,Hinged2,alambeigi2020steerable}. In particular,  we have introduced a unique Concentric Tube Steerable Drilling Robot  (CT-SDR) for creating C-, J- and U-shape trajectories and a complementary flexible pedicle screw for spinal fixation procedures \cite{Sharma2023ICRA,Sharma2024SpatialSF,Sharma2023TBME, Sharma2024TBME,Kulkarni2024TowardsTF, Maroufi2025S3D,Kulkarni2025SynergisticPSA,Kulkarni2025AugmentedBSF}. Despite CT-SDR's features, it can only be used for spine applications and creating short trajectories with one curvature, hindering its application for pelvic fixation where a long  and S-shape drilling trajectory with more than one curvature is required for placement of a long fixation implant in the complex anatomy of pelvis (see Fig. \ref{fig:Anat}-C).

To  address the   limitations of our current CT-SDR and as our main contribution, this paper introduces the design, fabrication, and evaluation of a novel four-degree-of-freedom CT-SDR for pelvic fixation procedures. This robot potentially enhances pelvic fixation surgery by accurately creating a nonlinear drilling trajectory    while respecting the   variable thickness and complex anatomy of the pelvis. As illustrated in Fig. \ref{fig:Anat}-C, the \textit{Pelvic CT-SDR} is designed with the goal of creating (i) long, (ii) planar and out-of-plane, (iii)  S-shaped drilling trajectories in the complex anatomy of the pelvis. The proposed modular design enables the usage of multiple pre-curved guiding tubes housing a flexible drilling instrument for creating various curved trajectories. 
 To  validate the functionality of the \textit{pelvic} CT-SDR with two pre-curved tubes, we performed  experiments to create different planar and out-of-plane S-shape drilling trajectories in simulated Sawbones phantoms.

\section{Design of \textit{Pelvic} CT-SDR System}
Ensuring that the \textit{pelvic} CT-SDR system can reliably create consistent S-shape curved trajectories is crucial for ensuring a safe procedure through the complex anatomical curvature of the pelvis. Similar to the design proposed in \cite{Sharma2023ICRA} and as shown in Fig. \ref{fig:Drill}, a multi-DoF \textit{pelvic} CT-SDR relies on 3 major components in order to create reliable smooth trajectories: (1) CT-SDRs utilize structurally strong yet superelastic concentric tubes to reliably steer the drill's tip to create complex trajectories (see Fig. \ref{fig:Drill}). The guide tubes of CT-SDR systems provide all of the steerable motion, with the interactions between their curvatures ``guiding" the system's tool tip through space. The \textit{spinal} CT-SDR presented in \cite{Sharma2023ICRA}, utilized a singular concentric guide tube in order to create 2 DoF motion, as the single tube could translate through a single curve or be rotated to remove a cavity. As a \textit{pelvic} CT-SDR, the system will need to create more complex S-shaped curvatures, requiring an additional concentric tube while matching the length requirements of pelvic screws (i.e., 90 mm in length \cite{Stevenson2017PercutaneousIF}); (2) To safely remove hard tissue, independent of the cutting angle and already cut trajectory, the CT-SDR contains a flexible cutting tool which can transmit rotational motion around corners and through straight paths (see Fig. \ref{fig:Drill}). These tools are further vital in ensuring that diameters drilled are approximately around 7 mm to ensure sufficient fixation \cite{Stevenson2017PercutaneousIF}; 
(3) And finally, to provide stability and control to the system, the main body of the CT-SDR comprises of the system's actuation unit containing motors and stabilizing frame. Of note, the proposed pelvic CT-SDR offers a modular design, allowing for the introduction of several nested  guide tubes (if needed). The designed system would also need to have a stiff structure to robustly handle the drilling forces exerted on the tubes and flexible tools and transfer it to the robot's structure. The following sections describe the main components of the \textit{pelvic} CT-SDR and how they satisfy the mentioned design requirements.

\subsection{Concentric Tubes}
To ensure that this system can create the required S-shaped trajectories, it is vital that the drill tip is safely guided to its desired location. Besides safely steering through the complex geometry of the pelvis,  this requires that the drilling length and diameter match with the existing drilling and fixation instruments (i.e., maximum drilling length and diameter of 90 mm and 7 mm, respectively \cite{Stevenson2017PercutaneousIF}).  To achieve this, our design leverages the superelastic properties of nitinol (Euroflex, GmbH, Germany). Nitinol (NiTi) is a biocompatible, superelastic shape memory alloy that can safely deform to a predetermined curvature, which is influenced by the heat treatment process applied. This heat treatment process is essential for ensuring our system can safely and reliably reach our chosen configuration with consistency. For this paper, without loss of generality and to prove functionality of our system, two NiTi tubes were used to create the planar and out-of-plane S-shaped trajectories. As shown in Fig. \ref{fig:Drill}, the outer tube has an outer diameter of 3.61 mm with a wall thickness of 0.25 mm and a length of 110 mm. The inner tube has an outer diameter of 2.6 mm with a wall thickness of 0.2 mm and a length of 290 mm. Both tubes were heat treated to a $k$=50 mm radius of curvature, following a similar procedure to those outlined in \cite{hodgson2001fabrication}. Of note, while the 50 mm radius of curvatures were arbitrarily chosen, they can be easily changed to match the patient's anatomy and the selected surgeon's  planned drilling trajectory. The smaller tube is nested within the larger tubes. The tubes are further housed inside a larger stainless steel tube, as shown in Fig. \ref{fig:Drill}, to ensure the NiTi tubes started in a straight configuration.

\begin{figure}[t]
    \centering
    \includegraphics[width=0.85\linewidth]{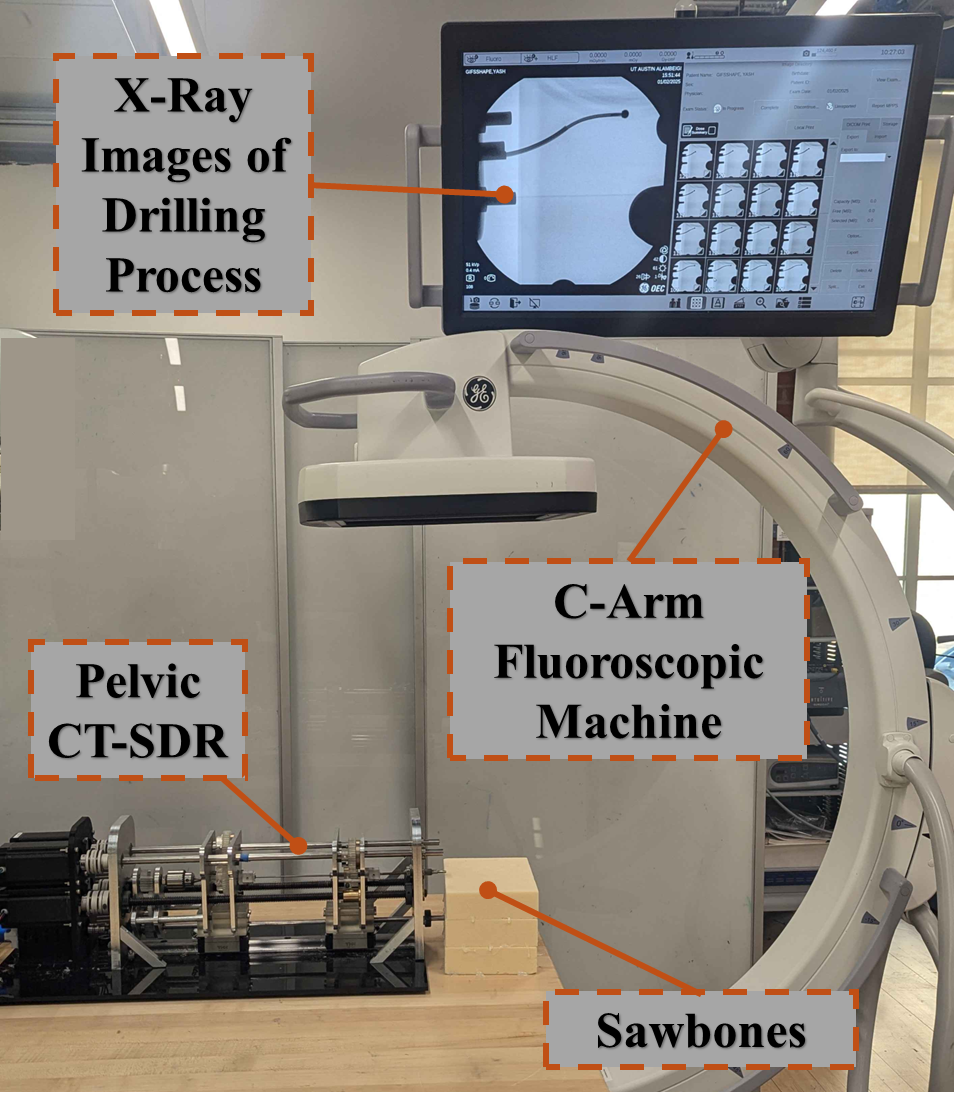}
     \caption{Experimental Set up utilized for evaluating the efficacy of the pelvic CT-SDR system, including a C-arm X-ray, pelvic CT-SDR system, and bone phantom.}
    \label{fig:ExperimentalSetup}
\end{figure}
\subsection{Cutting tool}
As shown in Fig. \ref{fig:Drill}, our flexible cutting tool system consists of three laser-welded components: (i) a ball nose end mill (42955A26, McMaster-Carr) with a 6 mm outer diameter (OD) used to create smooth curved trajectories while ensuring even material removal in all directions; (ii) A flexible torque coil with a 1.63 mm OD (Asahi Intec, USA, Inc.) is embedded within the curved NiTi tubes and reliably transmits rotational motion from the drill motor (B0C9956HMK, Amazon) to the drill bit; and  (iii), a 0.95 mm OD stainless steel tube is attached to the flexible torque coil, providing a stable interface between the drill chuck and the torque coil for safe and reliable operation.

\begin{figure*}[t]
    \centering
    \includegraphics[width=1\linewidth]{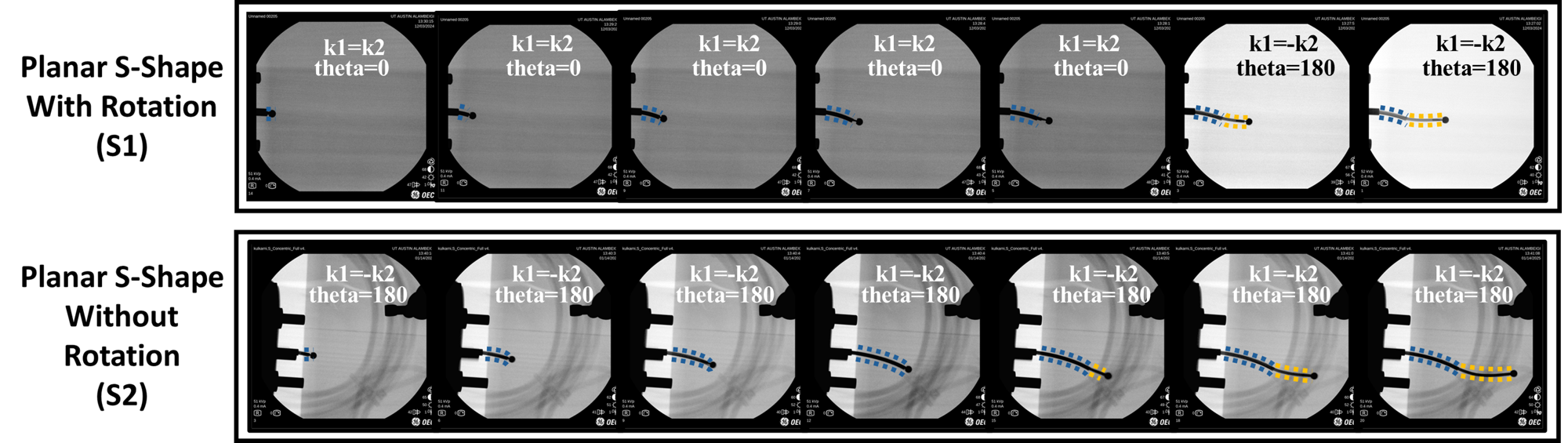}
     \caption{Fluoroscopic images illustrating two different scenarios of S-shaped planar drilling by the \textit{pelvic} CT-SDR. (Top) Planar S-shape drilling with rotation matching the description of scenario 1. (Bottom) Planar S-shape drilling without rotation matching the description of scenario 2. The path taken by the outer NiTi tube is marked in blue while the path of the inner NiTi tube is marked in yellow.  }
    \label{fig:Drilling}
\end{figure*}

\subsection{Actuation Mechanism}
In order to accurately control the location of the drill bit so that it can reach any desired configuration, it is essential to independently control the two  translational and two rotational degrees of freedom (DoFs) of the concentric tubes. To achieve this, two distinct mechanisms were implemented for controlling the translational and rotational motion. Figure \ref{fig:Drill} illustrates how a NEMA 23 stepper motor (6627T530,, McMaster-Carr) is connected to a lead screw (98940A305, McMaster-Carr) to control the translational movement for both inner and outer tubes. The NEMA 23 stepper motor is used to rotate a lead screw which in turn allowed the lead screw nut (6350K41, McMaster-Carr) to convert the rotational movement to a translation movement to move the NiTi Housing Unit illustrated in Fig.\ref{fig:Drill}-(1). To ensure smooth and precise movement, the NiTi Housing Unit was securely connected to a 3D-printed component and mounted on a linear rail (6709K431, McMaster-Carr), facilitating stable and controlled translational motion. 
The rotational movement of the concentric tubes is controlled by the NEMA 23 stepper motor connected to a spline shaft (61145K145, McMaster-Carr) also passing through the NiTi Housing Unit. This separate connection between the two systems allows for free translational motion while maintaining unimpeded rotational movement. The ball spline is secured within a rotational gear pair system with the system connecting the ball spline rotation to the rotation of the NiTi tubes. The gear system used is in a 1:1 ratio. This enables easy and consistent control of the gears by the motors.

Furthermore, the entire cutting tool system is driven by a drill motor shown in Fig. \ref{fig:Drill}. The drill motor is attached at the back of the system under the NEMA stepper motors and is attached to another ball spline that traverses the length of the entire system. This ball spline system is critical in allowing the drill carriage subsystem to move in unison with the translational motion of the inner tube NiTi housing system. The ball spline is linked to a gear system shown in the Drill Carriage subsystem shown in Fig. \ref{fig:Drill}-(2). This gear system works in a 1:1 ratio allowing the rotation of the drill chuck (2812A19, McMaster-Carr), which, in turn, drives the rotation of the drill bit. 
The NEMA 23 motors in this system were controlled using Rtelligent R60 motor drives (B07SBFZ596, Amazon). An Arduino Nano, running a custom-written Arduino program, managed the insertion and rotational mechanisms. The system's movement was controlled via joysticks, as shown in Fig. \ref{fig:Drill}-(3). Additionally, all components were integrated onto a custom-designed PCB board, as illustrated in Fig. \ref{fig:Drill}. 

To provide the stability required by the system, each carriage/housing unit and end plate are CNC cut from 0.25 in and 0.5 in, respectively, aluminum plates. This prevents deflections under the forces created by the interacting concentric tubes. Also to this purpose, external supports made from 0.25 in aluminum bars were added between the end plates and the table for additional support.

\section{Experimental Set-up and Results}

\begin{figure}[t]
    \centering
    \includegraphics[width=1\linewidth]{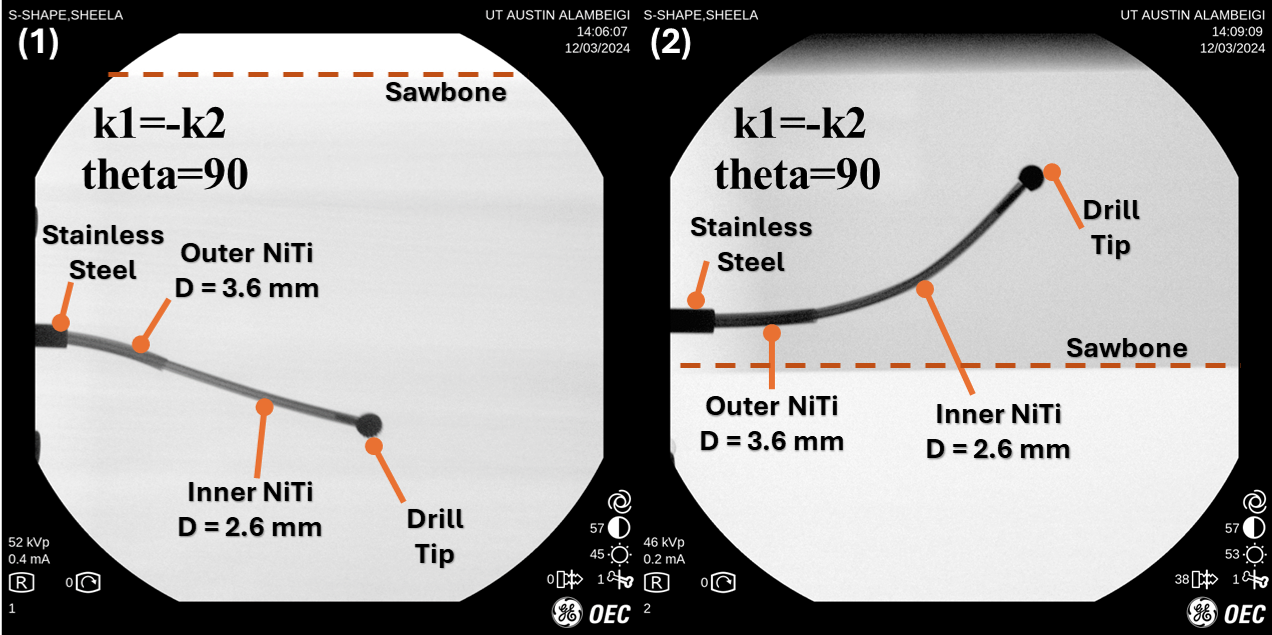}
     \caption{Out-of-Plane S-shaped drilling experiments following \textit{S1 scenario} with  $\theta=90$$\degree$. X-ray image illustrating S-shaped drilling from (1)  top  and (2)  from side viewpoints.}
    \label{fig:OutOfPlane}
\end{figure}

Figure \ref{fig:ExperimentalSetup} shows the experimental setup used to conduct experiments to validate the \textit{pelvic} CT-SDR’s ability to create planar and out-of-plane S-shaped trajectories. In addition to the \textit{pelvic} CT-SDR, the setup included a C-arm X-ray machine (OEC ONE CFD, GE Healthcare) to monitor the \textit{pelvic} CT-SDR's progress throughout the drilling procedure. The phantoms used during the procedures were made from PCF 5 Sawbones blocks (Pacific Research Laboratories, USA) as they closely resemble a similar density to osteoporotic bone \cite{etin2021ExperimentalIO}.
To demonstrate the ability of the pelvic CT-SDR system to create planar and out-of-plane trajectories, we performed the following experiments:
\subsection{S-shape planar  Drilling Experiments Under X-ray}  
For this study, we considered two different configurations for the curvature alignment of the nested tubes and their insertion length to evaluate  robustness of the robot versus the sequence of tubes insertions and rotations:

\subsubsection{Scenario 1 (S1)} As shown in the top row of Fig. \ref{fig:Drilling},
the NiTi tubes were first nested with their curvatures $k$ aligning together and fully constrained within the stainless steel outermost tube (i.e., initially $k_1=k_2$, $\theta=0$$\degree$). This allowed for a minimal amount of interaction between the two NiTi tubes during the first phase of their insertion. The \textit{pelvic} CT-SDR was then aligned with a Sawbones phantom, prior to the drill tip accelerating to 1000 rpm. Both NiTi tubes were then inserted simultaneously with about 20 mm length of the outer tube  to create the first section of the S-shaped trajectory. Next, the outer NiTi tube was held fixed as the inner NiTi tube was rotated $\theta=180$$\degree$ while it was  still constrained within the outer NiTi tube (i.e., final configuration: $k_1=-k_2$, $\theta=180$$\degree$). Of note, in this situation, the inner tube had the opposite curvature alignment with the outer tube.  The innermost tube was then translated forward, creating an opposed curved pathway to complete the S-shape path. This process was recorded with the C-arm X-ray and the progression of the test is shown as \textit{S1} in Fig. \ref{fig:Drilling}. 

\subsubsection{Scenario 2 (S2)} As shown in the bottom row of Fig. \ref{fig:Drilling},
the NiTi tubes were first nested with their curvatures aligning in the opposite direction of each other from the very beginning (i.e., initially $k_1=-k_2$, $\theta=180$$\degree$)  and fully constrained within the stainless steel outermost tube.  Both NiTi tubes were then inserted simultaneously with about 40 mm length of the outer tube  to create the first section of the S-shaped trajectory. Next, the outer NiTi tube was held fixed and the inner tube with opposite curvature  was then translated forward, creating an opposed curved pathway to complete the S-shape path (i.e., final configuration: $k_1=-k_2$, $\theta=180$$\degree$).  

\subsection{S-shape out-of-plane  Drilling Experiments Under X-ray}
The out-of-plane drilling trajectories, as shown in Fig. \ref{fig:OutOfPlane}, followed a similar procedure as the \textit{S1 scenario}. However, in order to guide the \textit{pelvic} CT-SDR's tip out-of-plane for the second half of the trajectory, the inner NiTi tube was rotated $\theta=90$$\degree$ prior to its translation out of the outer NiTi tube (i.e., final configuration: $k_1=-k_2$, $\theta=90$$\degree$). This experiment clearly shows the ability of pelvic CT-SDR in drilling out-of-plane trajectories orthogonal to each other. 

\subsubsection{Cross-Section and Repeatability analysis of S-shape  Drilling Trajectories}
 To closely simulate a realistic pelvic surgical fixation scenario, in this set of experiments we conducted repeated S-shape drilling tests with the drilling and fixation  parameters matching with the existing surgical approaches (i.e., maximum  drilling length of 90 mm with maximum hole diameter of 7 mm) \cite{Stevenson2017PercutaneousIF}.
 Therefore, for these experiments, we selected an outer tube arc length of approximately 40.7 mm and an inner tube arc length of around 50 mm. This allowed the outer tube to approach its maximum travel arc length (approximately 41.5 mm) while ensuring that the total length of the S-shaped curved trajectory match with a typical 90 mm screw size used in percutaneous sacroiliac screw fixation \cite{Stevenson2017PercutaneousIF}. Without loss of generality, we used constant radius of curvature of 50 mm  for both tubes. Other curvatures can readily be heat treated based on the planned trajectories.  
 
\begin{figure}[t]
    \centering
     \includegraphics[width=0.8\linewidth]{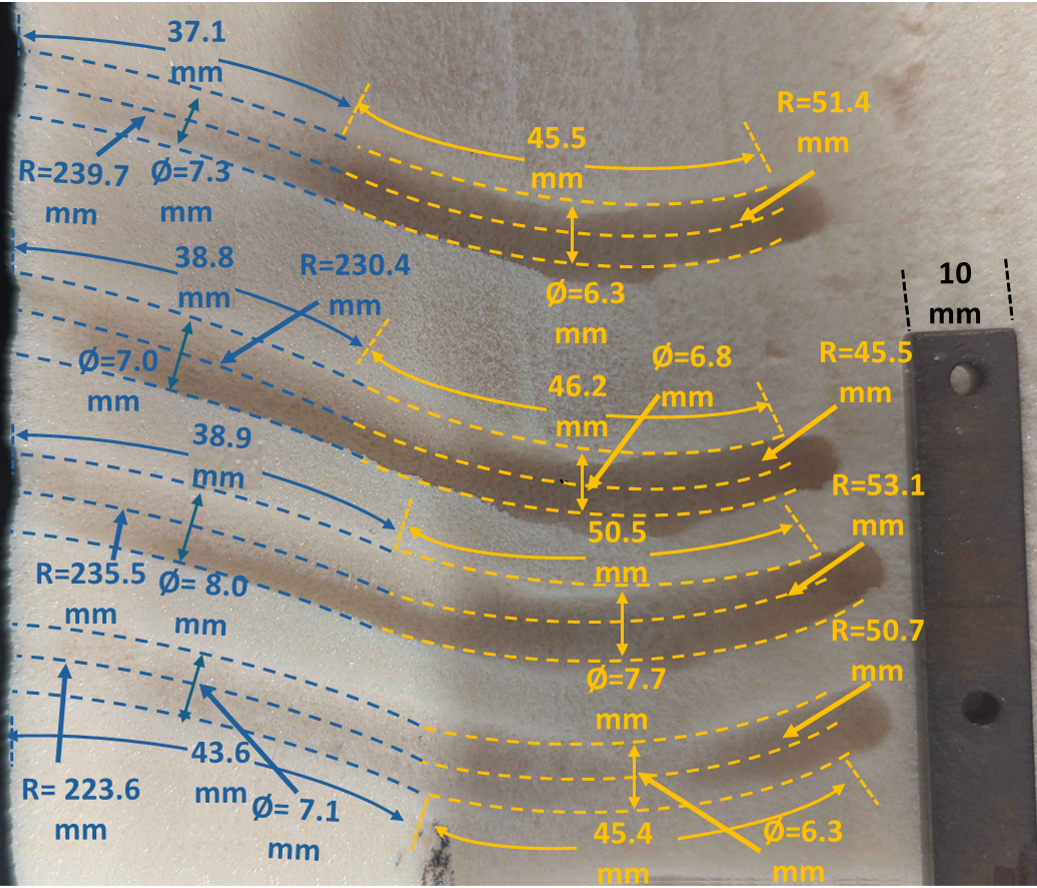}
     \caption{Experimental results indicating the insertion lengths, radius of curvature, and drilled diameter for both outer and inner NiTi drilled trajectories for S2 scenario.}
    \label{fig:Measurements}
\end{figure}

 To conduct this set of drilling  experiments, the \textit{pelvic} CT-SDR assembly was followed the \textit{S2 scenario } such that the tubes curvature initially opposed each other. Before insertion, the inner and outer NiTi tubes were concentrically nested, extending approximately 10.8 mm beyond the straight stainless steel tube to ensure sufficient clearance with the \textit{pelvic} CT-SDR system. The drill tip was aligned tangentially to the sample surface. The inner and outer NiTi tubes were inserted  simultaneously together to create the first section of the S-shape trajectory. Then,  the inner NiTi tube was solely advanced to form the complete S-shaped trajectory. After completion, the NiTi tubes were retracted, and the Sawbones block was cut in half. This procedure was repeated four times to assess repeatability of the drilling process. Next, as illustrated in Fig. \ref{fig:Measurements}, the cross section pictures of the S-shape trajectories were inserted in a 3D CAD software (SolidWorks) for further image processing and quantitative evaluation of the created tunnels.  Figure \ref{fig:Drilling} illustrate the arc lengths drilled as \textit{Planar S-shaped without Rotation (S2)} while the final experimental results are presented in Fig. \ref{fig:Measurements} and Table \ref{table1:Results}.

\section{Discussion}
With the goal of performing a steerable drilling process for  pelvic fixation procedures, we modified the design of our previously developed CT-SDR for spinal fixation applications \cite{Sharma2023ICRA, Sharma2023TBME}. Despite spinal fixation procedures requiring one straight and one curved tube to create a J-shape trajectory with maximum 50 mm length, to successfully perform a generic pelvic fixation procedure, the \textit{pelvic} CT-SDR needed to create long (i.e., approximately 90 mm) and S-shape trajectories. The efficacy of the \textit{pelvic} CT-SDR  is demonstrated by its ability in creating different planar and out-of-plane drilling trajectories as illustrated by the S1 and S2 scenarios shown in Fig. \ref{fig:Drilling}, Fig. \ref{fig:OutOfPlane}, and Fig. \ref{fig:Measurements}. In particular, Fig. \ref{fig:Drilling} highlights the \textit{pelvic} CT-SDR progression through two distinct S1 and S2 configurations. These X-ray snapshots clearly demonstrate the \textit{pelvic} CT-SDR's ability to create planar smooth S-shaped trajectories  with and without pre-rotation of the inner tube (i.e., different initial curvature alignments of the tubes). Further, Fig. \ref{fig:OutOfPlane} demonstrates the \textit{pelvic} CT-SDR's ability to drill out-of-plane trajectories inside a Sawbone phantom, with a 90$\degree$ offset from one another. Of note, this extreme drilling scenario, clearly validates the CT-SDR's ability in creating complex drilling trajectories.

\begin{table}[t]
%\begin{threeparttable}
\begin{center} 
\caption{Averaged Cross Sectional Measurements Shown in Fig\ref{fig:Measurements}. }
\label{table1:Results}
\setlength\tabcolsep{0pt} % make LaTeX figure out intercolumn spacing
\begin{tabular*}{0.75\columnwidth}{@{\extracolsep{\fill}} l cccc}
\Xhline{1.25pt}
\makecell{Tube} & \makecell{Inner+Outer \\ (mm)} & \makecell{Inner \\ (mm)}  \\
\Xhline{1pt}
\makecell{Ideal \\ Insertion \\ Length } & 40.7  & 50  \\
\Xhline{0.25pt}
\makecell{Measured \\ Insertion \\  Length } & 39.6 $\pm$ 2.4  & 46.9$ \pm$ 2.1  \\
\Xhline{0.25pt}
\makecell{Insertion \\Length\\Error} & 2.7\% & 6.2\% \\
\Xhline{0.25pt}
\makecell{Ideal \\ Radius of \\ Curvature } & N/A & 50 \\
\Xhline{0.25pt}
\makecell{Measured \\ Radius of \\ Curvature} & 232.3 $\pm$ 6.0 & 50.2 $\pm$ 2.9  \\
\Xhline{0.25pt}
\makecell{Radius of\\ Curvature \\ Error} & N/A  & 0.4\% \\
\Xhline{0.25pt}
\makecell{Drilled \\ Diameter} & 7.4  &  6.8  \\

\Xhline{1.25pt}
\end{tabular*}
\end{center}
%\end{threeparttable}
\end{table}

Figure \ref{fig:Measurements} and Table \ref{table1:Results} illustrate the results of the cross section analysis performed to show the reliability and repeatability of the drilled trajectories. From this analysis, we obtained an average measured insertion arc length of outer tube of 39.6 mm compared to the expected of 40.7 mm with an error of 2.7\%. For inner tube, we see a average insertion arc length of inner tube of 46.9 mm with an expected of 50 mm resulting in an error of 6.2\%. These low errors illustrate the ability of system to reliably create desired curved trajectories. Furthermore, the  inner tube radius of curvature has an average radius of curvature of 50.2 mm compared to the ideal heat treated radius of  50 mm with a low error of 0.4\%. This low error further verifies the pelvic CT-SDR's capability of creating reliable curvatures regardless of its interaction and drilling through Sawbones phantoms. Of note, in Table \ref{table1:Results}, we have not compared the expected and measured curvatures for the first section of the drilled S-shape trajectories. This is directly related to the  change in the overall curvature of inserted tubes compared with their initial heat-treated curvatures. Nevertheless, as expected in these experiments following the S2 drilling scenario with opposite insertion of curvatures, the overall measured curvature of the inserted tubes is about 4 times less than their initial  curvature (i.e.,  232 mm versus 50 mm). Furthermore, we measured an average drilled diameter of 7.4 mm for the outer tube and 6.8 mm drilled diameter for the inner tube, with an average drilled diameter of 7.1 mm. We measured an average insertion drilling time of 55 seconds. 

\section{Conclusion and Future Work}
With the goal of addressing the current limitations of  pelvic fixation procedures, caused by the rigidity of existing drilling and fixation instruments and complex anatomy of pelvis, we  proposed a novel four DoF \textit{pelvic} CT-SDR system to enable drilling planar and out-of-plane S-shape trajectories. For the first time, to our knowledge, we have (i) introduced the design and fabrication of this unique steerable drilling robot, and (ii) created planar  and out-of-plane S-shaped trajectories with different drilling scenarios. 

While this work has shown great promise, certain limitations remain to be addressed in our future studies. For example, in the future, we will focus on a thorough evaluation and assessment of this system considering different drilling configurations in animal bones and cadaveric specimens, incorporating both inner and outer tube insertion and rotations. Additionally, we will work on modeling the deformation behavior of the pelvic CT-SDR while drilling a hard tissue, which is necessary for precise S-shape trajectory following  and control. Our efforts will also focus on trajectory planning for precise image-guided control of the robot in complex anatomy of pelvis. 
 
\bibliographystyle{IEEEtran}
\bibliography{root}

\end{document}